\pdfoutput=1

\documentclass[11pt]{article}
\usepackage{stmaryrd}

\usepackage[final]{acl}

\usepackage{times}
\usepackage{latexsym}

\usepackage[T1]{fontenc}

\usepackage[utf8]{inputenc}

\usepackage{microtype}

\usepackage{inconsolata}

\usepackage{graphicx}
\usepackage{booktabs}
\usepackage{multirow}
\usepackage{caption}
\usepackage{subcaption}
\usepackage{tikz}
\usepackage{tabulary}
\usepackage{tabularray}

\usepackage[most]{tcolorbox}
\usepackage{listings}
\usepackage{fvextra}
\usepackage{framed}
\usepackage{comment}

\title{
Disambiguate First, Parse Later: Generating Interpretations for \\Ambiguity Resolution in Semantic Parsing
}

\author{Irina Saparina  \and Mirella Lapata \\  
  Institute for Language, Cognition and Computation,  \\
  School of Informatics, 
  University of Edinburgh  \\
  10 Crichton Street, Edinburgh EH8 9AB\\
    \texttt{i.saparina@sms.ed.ac.uk} \qquad  \texttt{mlap@inf.ed.ac.uk}}

\begin{document}
\maketitle
\begin{abstract}
Handling ambiguity and underspecification is an important challenge in
natural language interfaces, particularly for tasks like text-to-SQL
semantic parsing.  We propose a modular approach\footnote{Our code and data can be downloaded from \href{https://github.com/saparina/disambiguate-then-parse}{github.com/saparina/disambiguate-then-parse}.} that
resolves ambiguity using  natural
language interpretations before mapping
these to logical forms (e.g., SQL queries).
Although LLMs excel at parsing \emph{unambiguous} utterances, they show
strong biases for \emph{ambiguous} ones, typically predicting only preferred
interpretations. We constructively exploit this bias to generate an
initial set of preferred disambiguations and then apply a specialized
infilling model to identify and generate missing interpretations. To
train the infilling model, we introduce an annotation method that uses
SQL execution to validate different meanings.  Our approach improves interpretation coverage and generalizes across
datasets with different annotation styles, database structures, and
ambiguity types.
\end{abstract}

\section{Introduction}

\begin{figure}[t]
\centering
\includegraphics[width=\linewidth]{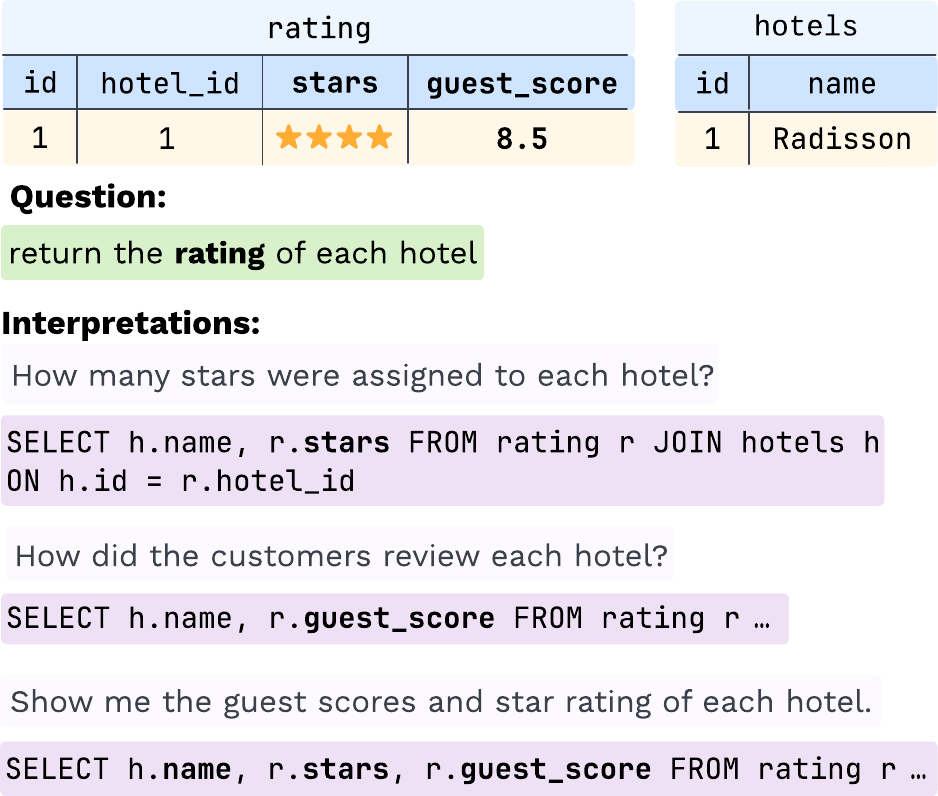}
\caption{Example of ambiguous question and   its  interpretations
  (in natural language and SQL).}
\label{fig:ambig_example}
\end{figure}

Natural language utterances are often ambiguous, vague, or
underspecified, giving rise to multiple valid interpretations. 
Figure~\ref{fig:ambig_example} shows an ambiguous request (``return
the rating of each hotel'') in the context of natural language
interfaces.  In the example, ``rating'' could refer to the number of
stars hotels receive as an indication of their quality (e.g.,~4 stars)
or guest reviews on booking websites (e.g.,~8.5 out of~10), or
both. Ignoring such ambiguity can lead to incomplete or incorrect
results, undermining user trust and limiting the practical usefulness
of any conversational system.

While state-of-the-art large language models (LLMs) demonstrate
remarkable performance on tasks like question answering,
text-to-SQL parsing, and natural language inference, recent studies
\citep{liu-etal-2023-afraid, Floratou2024} have shown they are not
adept at handling ambiguity. They  display systematic biases in their choice of interpretation \citep{kamath2024scope,
  eskin-2024-zero, Ambrosia}, typically defaulting to a single
interpretation when multiple ones exist. 

What are the response strategies LLMs should adopt to address
ambiguity? 
An approach might be to respond with a clarification question, which
directly engages users and ensures accurate resolution of the
ambiguity but introduces additional interaction turns. Alternatively,
presenting multiple possible interpretations (an ``overton response'')
would allow users to select the most relevant answer themselves.  This
approach minimizes interruptions, caters to users with different
levels of expertise, and provides interpretability by making the
system's reasoning explicit.  For tasks like semantic parsing (see
Figure~\ref{fig:ambig_example}), a hypothetical response should not
only include interpretations in their final form (e.g.,~SQL queries)
but also their different readings in natural language. From a modeling
perspective, interpretations  serve as
explanations (of the system's output) and as \emph{intermediate
representations}, providing a way to decompose complex semantic
parsing problems into simpler steps.

\begin{figure*}[t!]
\centering
\includegraphics[width=0.98\linewidth]{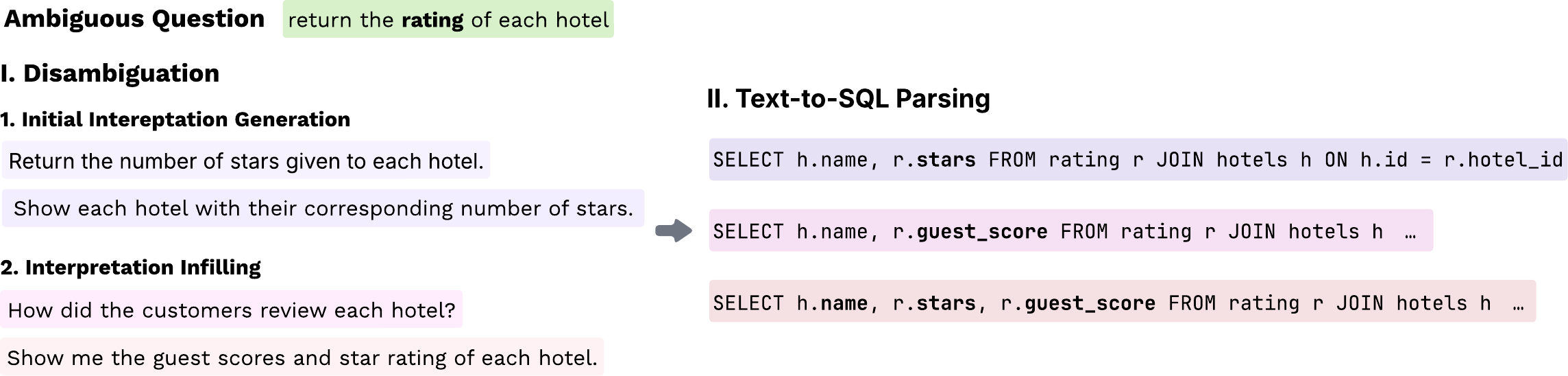}
\caption{Disambiguate first parse later: we disambiguate database
  questions into natural language interpretations (by generating an initial
  set and filling in missing ones), and  then a  parser translates each
   interpretation into SQL.}
\label{fig:approach}
\end{figure*}

\looseness=-1
In this work, we focus on text-to-SQL parsing of ambiguous questions,
and propose a two-stage approach that first disambiguates by generating
all possible meanings in natural language, and then parses each
unambiguous interpretation. Separating the disambiguation and parsing
tasks allows us to use existing semantic parsers that perform
generally well on \emph{unambiguous} inputs. We obtain
interpretations, by prompting an LLM to generate all possible meanings
for an utterance. These initial interpretations are often incomplete
due to inherent biases stemming from statistical patterns found in LLM
training data, lack of intrinsic knowledge, and different alignment
preferences. Rather than attempting to correct these biases, we
introduce an infilling model that reviews the ambiguous
question and initial interpretations, and then generates any missing
ones.

\looseness=-1
The infilling model is trained on pairs of default interpretations
and missing readings. Problematically, existing  datasets 
provide (multiple) SQL parses for ambiguous questions without explicitly
verbalising the interpretations they correspond to. We create
synthetic reference interpretations for AmbiQT
\citep{bhaskar-etal-2023-benchmarking}, a recently proposed benchmark
for parsing ambiguous questions into SQL. We exploit the
fact that generated interpretations can be converted to SQL queries
which we execute to verify whether they are correct \emph{and} to
establish which interpretations are missing, i.e.,~whether there exist
gold SQL parses for which no interpretation was found. We evaluate our
approach on Ambrosia \citep{Ambrosia}, a dataset different from AmbiQT
in terms of database content and the types of ambiguity it
represents.  Our contributions
are summarized as follows:

\begin{itemize}
  \itemsep0em 
\item  We propose a modular approach that uses natural language
 to spell out ambiguity before mapping individual interpretations
 to logical forms (e.g., SQL queries).
\item We use LLMs to generate an initial set of preferred disambiguations
and then apply a specialized infilling 
model to identify and generate missing interpretations.
\item Experiments show that our ``disambiguate first
parse later'' strategy improves the coverage of interpretations for
ambiguous questions  and generalizes across annotation
styles, database structures, and ambiguity types.

  \end{itemize}

\section{Disambiguate First Parse Later}

\subsection{Problem Formulation} 
Semantic parsing is the task of mapping a natural language
utterance~$u$ to  formal expression~$e$ in grammar~$G$, where~$e$
captures the intended meaning of~$u$. The expression $e$ can be then
executed in an environment~$\mathcal{E}$ to produce a
denotation~$\llbracket e \rrbracket$. In the unambiguous case, there
is a single valid expression~$e$ that corresponds to the user's
intent. However, when $u$~is ambiguous, there are multiple valid
expressions $\{e_1, ..., e_n\}$ where $\llbracket e_i \rrbracket \neq
\llbracket e_j \rrbracket$ for some $i,j$, with each $e_i$ representing a
valid interpretation of the user's intent.

In our text-to-SQL task, grammar~$G$ defines valid SQL queries for a
given database schema~$S$. The database schema and table descriptions provide
context~$\mathcal{C}$ that can help disambiguate some queries. We
focus on questions that remain ambiguous even when $\mathcal{C}$~is
known, i.e.,~there exist multiple valid SQL queries $\{e_1, ...,
e_n\}$ that respect~$S$ and produce different result sets $\llbracket
e_i \rrbracket$ when executed on the database. In the example in
Figure~\ref{fig:ambig_example}, the question $u$ = ``return the rating
of each hotel'' remains ambiguous for a given schema $S$ (see top
table), as ``rating'' could refer to star ratings and guest reviews
even when the database content is known.

\subsection{Natural Language Interpretations for Explicit Disambiguation}

We propose to resolve
ambiguity prior to generating SQL expressions with natural language interpretations.  More formally, for 
ambiguous utterance~$u$, we first produce a set of unambiguous natural
language interpretations $\{\hat{u}_1, ..., \hat{u}_n\}$, where each~$\hat{u}_i$ captures one  meaning of~$u$. We then
map each~$\hat{u}_i$ to formal expression~$e_i$. %
For example, given $u$ = ``return the rating of each
hotel'', our goal is to produce $\hat{u}_1$ = ``How many stars were
assigned to each hotel?'', $\hat{u}_2$ = ``How did the customers
review each hotel?'' and $\hat{u}_3$ = ``Show me the guest scores and star rating of each hotel'' and map them to corresponding SQL queries $e_1$, $e_2$ and $e_3$
(see Figure~\ref{fig:ambig_example}). This approach has several advantages:
\begin{itemize}
    \itemsep0em 
  \item Unambiguous interpretations $\hat{u}_i$ are easier to translate into
    formal expressions, as they can be handled by standard semantic
    parsers (e.g.,~existing text-to-SQL models);
   \item Natural language interpretations provide transparency, making the system's internal working explicit to users;
   \item The modular design allows us to optimize the disambiguation
     and formal expression mapping components independently.
\end{itemize}

Building on this modular approach, we propose to further decompose
explicit disambiguation into two steps: initial interpretation
generation followed by infilling of missing interpretations. Our core
idea is based on the observation that \textbf{modern LLMs, like
  humans, resolve ambiguous questions by gravitating toward preferred,
  default interpretations} \citep{kamath2024scope, eskin-2024-zero,
  saparina-lapata-2024-improving}. We leverage this tendency by using
LLMs to generate preferred interpretations. We then train a
specialized model that, given an ambiguous question and its default
interpretations, identifies and generates missing readings to ensure
all valid meanings are covered.  Figure~\ref{fig:approach} illustrates
our approach: generating default interpretations
(using LLMs), infilling missing ones, and text-to-SQL parsing.

 \subsection{Default Interpretation Generation}
 LLMs exhibit strong biases in interpretation generation,
consistently favouring certain types while missing others. 
We  take advantage of these biases by 
using LLMs to generate an
initial set of interpretations for ambiguous utterances.

\looseness=-1
The first component of our approach is a prompt-based interpretation
generator that produces candidate interpretations given utterance~$u$
and database context~$\mathcal{C}$. We prompt an LLM to identify and
list all \textit{semantically different} interpretations of $u$. 
In practice, LLMs generate only a subset of possible interpretations, typically the most straightforward ones. We refer to them as \textbf{preferred} or \textbf{default interpretations}.
The
prompt is designed to handle \emph{both} ambiguous and unambiguous
cases, allowing the model to return a single interpretation when
appropriate. As we rely on the LLM's ability to disambiguate to its
default interpretations, no additional training is required. The output of this module is a set of natural language
interpretations $\{\hat{u}_1, ..., \hat{u}_k\}$, which are mapped
to formal expressions $\{e_1, ..., e_k\}$ using a standard semantic
parser. For text-to-SQL tasks, executing these queries allows us to
identify paraphrases (interpretations that lead to identical execution
results) and filter redundancy.

\begin{figure*}[t]
\centering
\includegraphics[width=\linewidth]{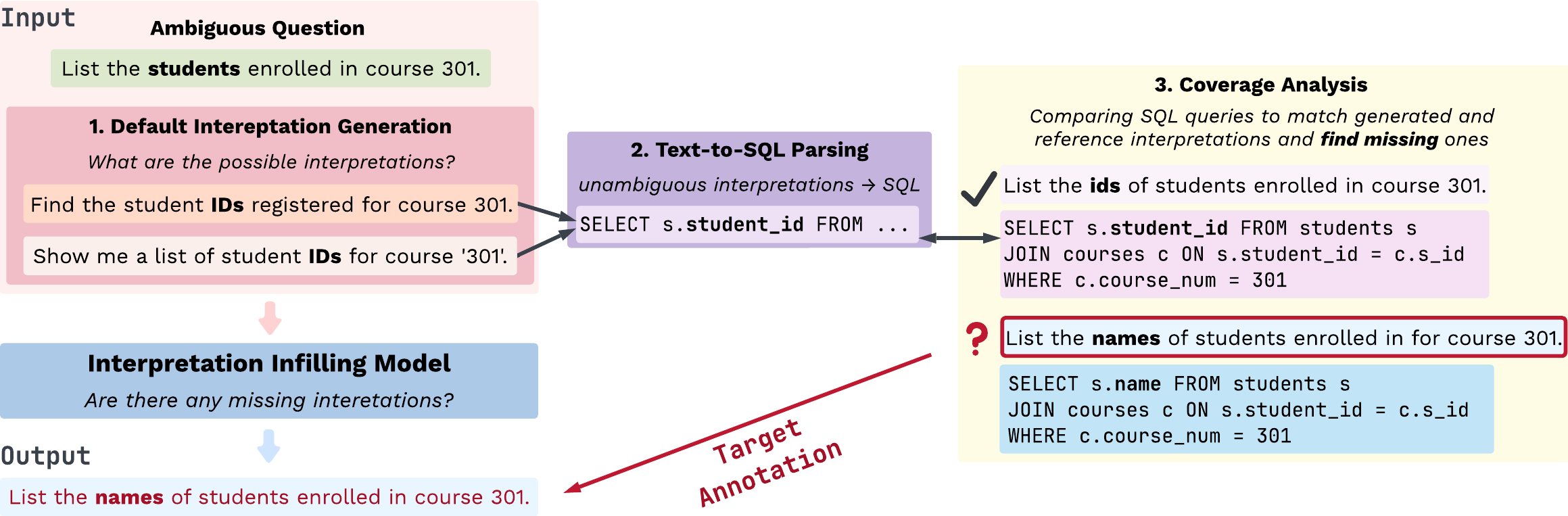}
\caption{The Infilling Model receives an ambiguous question and
  initial default interpretations as input and generates missing
  interpretations as output. Supervision comes from comparing SQL
  queries for default interpretations with gold SQL queries to
  identify which disambiguations are not captured.}
\label{fig:method_annotation}
\end{figure*}

\subsection{Interpretation Infilling}\label{sec:infilling_model}
While initial interpretations from LLMs capture a common understanding
of queries, they often miss valid alternative meanings.  We propose to
address this issue by introducing a specialised model that identifies
and generates missing interpretations.

Given  utterance~$u$, database context~$\mathcal{C}$, and
default interpretations $\{\hat{u}_1, ..., \hat{u}_k\}$, our model
determines whether additional interpretations exist and generates
these accordingly.  The output is a set of 
interpretations $\{\hat{u}_l, ..., \hat{u}_m\}$ which complement the
initial set. These interpretations are then also mapped to formal
expressions $\{e_l, ..., e_m\}$ (SQL queries in our case)
using a standard semantic parser.
We train the infilling model in a supervised manner. For now, let us assume we
have access to reference interpretations and corresponding gold SQL
queries.  By comparing the set of default interpretations with
reference interpretations, we identify which meanings are missing from
the initial set. Reference interpretations absent from the default set
serve as target outputs for the infilling model.

Determining whether two sentences have the \emph{same} meaning, given
some (database) context, is an extremely challenging
task. However, instead of comparing natural language expressions
directly, we compare their corresponding SQL queries. In particular,
we know which gold SQL queries are associated with each reference
interpretation. Similarly, we can obtain SQL queries predicted by a
text-to-SQL parser for the initial set of default
interpretations.
By executing these queries and comparing their results, we determine
which interpretations match and are thus covered by the initial
set. Reference interpretations absent from this set serve as
training targets for the infilling model which also  explicitly indicates when all interpretations are covered by the
initial set, and there is nothing to add (the target output in this
case is the sentence: ``All interpretations are covered.'')
Figure~\ref{fig:method_annotation} illustrates the annotation process
just described.

\subsection{Discussion}
The strength of our proposal lies in its modular design, consisting of
three separate components: default interpretation generation,
infilling, and text-to-SQL parsing. Only infilling requires training,
since we resort to existing pre-trained models for generating
interpretations and parsing these into SQL. This plug-and-play
functionality allows us to experiment with different LLMs 
and text-to-SQL parsers, without
retraining. Moreover, any improvements in the infilling module should
in theory translate into improved coverage.

A potential limitation is that infilling requires reference
interpretations and SQL queries, which may be difficult to come by. In
the next section, we demonstrate how reference interpretations can be
synthetically generated. In addition, the text-to-SQL parser may
occasionally introduce errors, leading to duplicate interpretations in
the final set, coming from the initial generator and infilling
model. However, this redundancy can improve robustness by
providing multiple formulations of the same meaning. Additionally,
duplicates can be filtered in post-processing by comparing execution
results. It is also important to note that the infilling does not
filter any interpretations from the initial set; it can only add new
ones or leave the set unchanged. This means that, in some cases,
incorrect interpretations may persist, potentially reducing
precision. However, we believe this is not a major concern in
practice, as users can simply ignore irrelevant
interpretations. %

\looseness=-1
When questions are unambiguous, the initial set contains a single
interpretation (or paraphrases with the same meaning, which can be
filtered as duplicates), and the infilling model  simply outputs
``All interpretations are covered.'' This means that we do not need to
explicitly determine whether a question is ambiguous or not, our
approach naturally handles both cases.

\section{Experimental Setup}

\subsection{Evaluation Datasets}
We evaluate our approach on two recent benchmarks aiming to assess
the capabilities of text-to-SQL parsers when faced with ambiguous
questions:  AmbiQT \citep{bhaskar-etal-2023-benchmarking} and Ambrosia
\citep{Ambrosia}. Examples can be found in Appendix~\ref{appx:dataset_examples}.

\paragraph{AmbiQT}\hspace*{-1.5ex}builds upon the widely used Spider dataset \citep{yu-etal-2018-spider}. Specifically, ambiguity is injected by generating synonyms
for column and table names (via ChatGPT and heuristically), by having
tables with overlapping column names (which leads to \texttt{join}
ambiguities), and by introducing columns which are aggregates of
certain values in addition to aggregating values with
the \texttt{group-by} clause of the \texttt{SELECT} statement. AmbiQT
inherits from Spider diverse SQL queries and
real-world databases, but also introduces redundancy as the databases
can contain duplicates (e.g., identical columns ``singer'' and
``performer'').  It  features two types of ambiguity, namely
lexical ambiguity (based on ambiguous column and table names) and
structural ambiguity in SQL queries (due to \texttt{join} and
pre-computed aggregates).
While the AmbiQT test set contains 3K examples, we filter
out those that yield empty execution results, leaving a final set of 1.8K~non-trivial examples.

\paragraph{Ambrosia}\hspace*{-1.5ex}showcases three different types of ambiguity,
namely scope ambiguity, attachment ambiguity, and vagueness.  It
contains human-written questions, SQL queries either manually written
or based on templates, and synthetically generated
databases. In addition to ambiguous questions and corresponding SQL
queries, Ambrosia has explicit interpretations in natural language.
The official test set has 1K~ambiguous questions and 2K
interpretations. We use the ambiguous subset in all experiments unless
otherwise stated. Notably, AmbiQT and Ambrosia handle vagueness differently:
AmbiQT consistently provides two valid interpretations, whereas
Ambrosia may provide up to three interpretations for vague questions.

\subsection{Evaluation Metrics}
\looseness=-1
Our primary metric is \textbf{Full Interpretation Coverage}, which measures the
proportion of examples where \emph{all} valid SQL interpretations are
present in the output. This metric aligns with the ``All Found''
metric from Ambrosia and ``BothInTopK'' from AmbiQT. 
Full Coverage is the most informative and relevant metric for our task, as it directly measures  the system’s ability to recover the full set of available interpretations. 
We also report \textbf{Single
  Interpretation Coverage}, i.e.,~the proportion of examples where at
\emph{least one} valid interpretation is found (``EitherInTopK'' from
AmbiQT),  as well as \textbf{Precision} and \textbf{Recall}.
We compare SQL queries based on their execution results.

\begin{table*}[t]
\centering
\begin{tabular}{@{}lcccccccc@{}}
\toprule
& \multicolumn{4}{c}{AmbiQT} & \multicolumn{4}{c}{Ambrosia} \\
\cmidrule(lr){2-5} \cmidrule(l){6-9}
\raisebox{.8em}[0pt]{\textit{End-to-End Text-to-SQL}}  & Single & Full & Recall & Precision & Single & Full & Recall & Precision \\
\midrule
0-shot Prompt         & 62.3  & 12.3  & 37.3   & 58.1   & 29.4  &  0.9  & 15.0   & 21.9   \\
3-shot Prompt         & 44.3  & 10.9  & 27.6      & 33.0      & 35.7  &  1.3  & 17.5      & 21.3      \\
SFT                   & 82.1  & \textbf{63.2} & 72.7   & 74.3   & 38.0  &  0.4  & 20.0   & 29.4   \\
\bottomrule
\multicolumn{9}{c}{~} \\ \toprule
& \multicolumn{4}{c}{AmbiQT} & \multicolumn{4}{c}{Ambrosia} \\
\cmidrule(lr){2-5} \cmidrule(l){6-9}
\raisebox{.8em}[0pt]{\textit{Disambiguate-and-Parse}}  & Single & Full & Recall & Precision & Single & Full & Recall & Precision \\
\midrule
Interp.\ Prompt       & 81.8  & 26.0  & 53.4   & 31.8   & 81.9  & 16.9  & 49.0   & 29.1   \\
w.\ Self-Correction   & 77.4  & 13.9  & 45.7      & 46.0      & 65.7  &  5.9  & 34.5     & 29.4     \\
Gold Interp.\ SFT     & 87.4  & 61.2  & \textbf{74.1} & \textbf{79.6} & 62.6  &  0.3  & 32.1   & \textbf{49.5} \\
Ours                  & \textbf{92.3} & 53.2  & 72.8   & 38.2   & \textbf{84.4} & \textbf{18.8} & \textbf{51.9} & 24.2   \\
\bottomrule
\end{tabular}
\caption{
\textbf{Single} and \textbf{Full} Interpretation Coverage, \textbf{Recall} and \textbf{Precision} (\%) on AmbiQT and Ambrosia datasets.
}
\label{tab:main_results}
\end{table*}

\subsection{Training Data}\label{sec:ambiqt_aug}
\looseness=-1
To train the infilling model, we need three components: (1)~generated
default interpretations, (2)~reference SQL queries, and (3)~reference
interpretations to identify missing interpretations  from the default
set. It is straightforward to elicit default interpretations from an
LLM and SQL queries are available in many text-to-SQL datasets. To
collect reference interpretations, we propose a novel approach that
extends the data generation process in AmbiQT
\citep{bhaskar-etal-2023-benchmarking}.

\looseness=-1
AmbiQT  relies on ChatGPT to generate two synonyms for a selected
column or table in an SQL query which naturally renders the
corresponding question vague. The original SQL mentions 
 are replaced with their synonyms, resulting in two gold SQL
queries. Building upon this approach, we use these synonyms and prompt
an LLM (with three in-context examples) to replace the original
mentions in the \emph{questions} with their synonyms. Our experiments use the
instruction tuned Llama 3.1~8B \citep{Dubey2024TheL3}. The full
prompt is provided in Appendix~\ref{appx:AmbiQT_syn_prompt}.  We
verify the quality of the synthetically generated interpretations by
attempting to generate correct SQL queries using a specialized code
generation LLM (instruction-tuned Qwen2.5-Coder~32B;
\citealt{hui2024qwen25codertechnicalreport}). We only accept examples where both
interpretations succeed within five attempts.

\looseness=-1
This approach is  particularly effective for AmbiQT and other datasets based on Spider  \citep{yu-etal-2018-spider} as
it contains many questions which directly mention table and column
names \citep{deng-etal-2021-structure, suhr-etal-2020-exploring,
  gan-etal-2021-towards}, making synonym substitution natural and
fail-safe.
We generate interpretations for approximately 5K~examples
from a subset of the Spider training data. Although our experiments
focus primarily on AmbiQT, the proposed interpretation generation
method can be applied to other domains and datasets (together with
AmbiQT's approach of generating ambiguous examples using column and
table synonyms).

\subsection{Implementation Details}

The first component of our method is to
generate default interpretations for an ambiguous question. We design a zero-shot prompt that
uses the provided database and question to generate interpretations
(see Appendix~\ref{appx:default_interp_prompt}). Here and throughout,
we represent the database as an SQL dump. For our experiments, we use
instruction-tuned Llama-3.1 8B, as it
produced the most coherent interpretations among similarly sized
models.

The second component is the infilling model which takes the database,
question, and default interpretations as input, and outputs
missing interpretations. As our infilling model, we train a LoRA
adapter \citep{lora-hu-etal-2022} on top of the instruction-tuned
Llama-3.1 8B (see Appendix~\ref{appx:infill_details}).

The final component is a text-to-SQL model for which we select the
instruction-tuned Qwen2.5-Coder-32B, a specialized model for
code generation. We also use this model to match and identify missing
interpretations (Section~\ref{sec:infilling_model}). A zero-shot
prompt for unambiguous text-to-SQL parsing is provided in
Appendix~\ref{appx:text2sql_prompt}.  Thanks to our modular structure,
any component of our system can be easily replaced with a more
powerful or more efficient model if needed.

\section{Experimental Results}

\paragraph{It is better to disambiguate first and parse later both in
  in-domain and out-of-domain settings.} Table~\ref{tab:main_results}
presents our main experimental results on AmbiQT and Ambrosia. We
compare our approach with both prompt-based and fine-tuning
baselines. Note that for methods requiring fine-tuning, AmbiQT
represents in-domain evaluation, whereas evaluation on Ambrosia is out of
domain. All methods use the same model, instruction-tuned
Llama-3.1 8B, and those that require training are fine-tuned using a LoRA adapter on the same AmbiQT subset, augmented with interpretations.

Table~\ref{tab:main_results} is split into two sections. The first one  presents end-to-end approaches, which attempt to \emph{directly} predict
multiple SQL queries for ambiguous questions. We report results for
zero-shot prompting, few-shot prompting, and end-to-end
fine-tuning (SFT). For prompting, we follow
\citet{Ambrosia} and explicitly instruct the model 
to generate multiple SQL
queries if the question is ambiguous. For few-shot prompting, we
sample 3~random examples from the corresponding dataset.  The second
section in Table~\ref{tab:main_results} lists methods which use
natural language interpretations to disambiguate first and then  
rely on text-to-SQL parsing of unambiguous questions. We report results for
generating all possible interpretations through LLM prompting
(which corresponds to our method without infilling), applying self-correction to this approach (details are in Appendix~\ref{appx:self_correction}), 
a fine-tuning
method which is trained to generate reference interpretations instead
of SQL queries, and our proposal which uses infilling to augment the
set of default interpretations. We use an instruction-tuned Qwen2.5-Coder 32B for text-to-SQL generation.

\looseness=-1
As can be seen in Table~\ref{tab:main_results}, prompting (0-shot,
3-shot) performs poorly on ambiguous questions, which is consistent
with the findings of \citet{Ambrosia}. Fine-tuning achieves excellent results on in-domain evaluation (AmbiQT), but fails to generalize to
Ambrosia, which suggests that the model overfits specific patterns in
AmbiQT. Interpretation generation (via prompting) shows promising
results on both datasets compared to end-to-end methods. 
Self-correction increases precision but also eliminates some valid interpretations, suggesting that the filtering task is non-trivial and requires further research.
Fine-tuning
on interpretations behaves very similarly to fine-tuning on SQL
queries, but provides higher precision and single-interpretation coverage, which
suggests it is more accurate in predicting \emph{at least one} correct
interpretation. 

\looseness=-1
Interpretation generation with infilling further improves single and
full interpretation coverage on both datasets.  While it does not
achieve the highest full coverage on AmbiQA, it substantially improves
over end-to-end prompting reaching~53\%, while delivering the best
single interpretation coverage of~92\%. It effectively generalises to
new domains and ambiguity types, showing the best results on Ambrosia
among all methods {across three out of four evaluation metrics}. However, the full coverage on Ambrosia is
still relatively low (at~19\%). Ambrosia is a challenging benchmark on its own
 but it is also possible that some interpretations are
missed due to annotation differences between the two datasets.

\looseness=-1
The results in Table~\ref{tab:main_results} show that recall closely mirrors full coverage, offering little additional insight. Precision, on the other hand, can be misleadingly inflated when systems rarely predict all valid interpretations at once. For this reason, in subsequent experiments, we only report  full and single interpretation coverage as the most informative metrics. See Appendix~\ref{appx:additional_results} for additional results.

\paragraph{Infilling boosts performance across interpretation
  generation models.}  We next analyze key components of our approach
through ablation studies. Table~\ref{tab:ablation_interp} focuses on
the first stage of our method, namely disambiguation. We compare
default interpretations from three different instruction-tuned models
of similar size: Qwen-2.5 7B \citep{Yang2024Qwen25TR}, Llama-3.1 8B, and
Gemma-2 9B \citep{Riviere2024Gemma2I}. We observe that Llama-3.1 8B
provides the best default interpretations and our infilling model
improves upon all interpretations, irrespective of the generation
model.

\begin{table}[t]
\centering
\begin{tabular}{@{}lc@{\;\;\;}cc@{\;\;\;}r@{}}
\toprule
& \multicolumn{2}{c}{AmbiQT} & \multicolumn{2}{c}{Ambrosia} \\
\cmidrule(lr){2-3} \cmidrule(l){4-5}
Interpretations & Single & Full & Single & Full \\ \midrule
Llama 3.1 8B & 81.8 & 26.0 & 81.9 & 16.9 \\
\hfill w. infilling &  92.3 & 53.2 & 84.4 & 18.8 \\
Qwen 2.5 7B   & 65.0 & 25.9 & 77.4  &  12.4 \\
\hfill w. infilling & 88.7 & 48.6 & 80.0 &  13.6\\
Gemma 2 9B   &  77.3 & 17.3  & 69.3 &  2.9 \\
\hfill w. infilling &  91.1 &  55.9 &  76.6 &  5.3 \\\midrule
Gold Interp.  & --- & --- &  75.2 &  49.0 \\ 
\bottomrule
\end{tabular}
\caption{\textbf{Single} and \textbf{Full} Interpretation Coverage
  (\%) on AmbiQT and Ambrosia when comparing different models for
  default interpretation generation, both with and without
  infilling. We also include an upper bound obtained using gold
  interpretations from Ambrosia.}
\label{tab:ablation_interp}
\end{table}

\begin{table}[t]
\centering
\begin{tabular}{@{}l@{\;\;\;}c@{\;\;\;}c@{~~}c@{\;\;\;}c@{}}
\toprule
& \multicolumn{2}{@{}c@{}}{AmbiQT} & \multicolumn{2}{c@{}}{Ambrosia} \\
\cmidrule(lr){2-3} \cmidrule(l){4-5}
Text-to-SQL Model & Single & Full & Single & Full \\ \midrule
  Qwen2.5-Coder 32B & 92.3 & 53.2 & 84.4 & 18.8 \\
  Qwen2.5-Coder 7B & 84.7 & 40.2 &  72.2 & 10.9 \\
\bottomrule
\end{tabular}
\caption{\textbf{Single} and \textbf{Full} Interpretation Coverage
  (\%) on AmbiQT and Ambrosia datasets when different text-to-SQL
  models are used.}
\label{tab:ablation_coder}
\end{table}

\begin{table}[t]
\centering
\begin{tabular}{@{}l@{\;\;\;}c@{\;\;\;}c@{}}
\toprule
Error Type & AmbiQT & Ambrosia\\
\midrule
Redundant interpr.& 23.3 & 20.0 \\
Overly general interpr. & 16.7 & 20.0 \\
Interpr. not matching intent & 26.7& 30.0 \\
SQL generation errors & 33.3 & 30.0  \\ 
\bottomrule
\end{tabular}
\caption{Error types (\%) on AmbiQT and Ambrosia.}
\label{tab:error_analysis}
\end{table}

\paragraph{Text-to-SQL parsing is hard even with gold interpretations!}
To provide an upper bound for our approach, we use gold
interpretations from Ambrosia and evaluate how well the text-to-SQL
parser performs when all interpretations are correct. The results are,
as expected, significantly higher, indicating room for improvement in
interpretation generation. However, since the full coverage reaches
only 49\%, a substantial number of errors may come from the
unambiguous text-to-SQL parsing alone.

Table~\ref{tab:ablation_coder} compares our text-to-SQL model, the
instruction-tuned Qwen-2.5 Coder 32B, with its smaller 7B
variant. Results decrease substantially when the weaker text-to-SQL
model is used. As our approach is modular, we anticipate our results
would improve with a better text-to-SQL parser.

\paragraph{Generalisation to new ambiguity types remains an open challenge.}
To better understand our system's limitations, we performed a manual error analysis on 50~randomly sampled examples with errors (30 from AmbiQT and 20 from Ambrosia). 
Table~\ref{tab:error_analysis} summarizes the distribution of these errors.  Redundant interpretations refer to near-paraphrases that differ slightly in structure. If their SQL queries yield the same result, we typically filter them out; however, minor variations (e.g., extra columns) can lead to execution mismatches, which are considered errors.
Additionally, some interpretations remain overly vague,  failing to fully disambiguate the input. These two error categories  mostly originate from default interpretations (~75\%). We also observed SQL  errors for valid interpretations, with~57\%  stemming from  infilled interpretations, including incorrect joins and column mismatches. Syntax errors occurred in up to ~20\% of cases.

 Most AmbiQT failures are due to join and aggregate ambiguities, which are not represented in the training data. Ambrosia presents even greater challenges with  scope and attachment ambiguities, which remain particularly difficult.

\looseness=-1
\paragraph{Disambiguation also improves coverage for unambiguous questions.}
While our approach does not specifically target unambiguous examples,
in Table~\ref{tab:ambrosia_unambig} we examine how different methods
perform when the input question is unambiguous, i.e.,~when it has one
interpretation and one SQL query. In this case, single and full
coverage are the same and we simply show whether the gold SQL query
was found. Note that we do not penalize additional queries in the
answer.

\looseness=-1
As Table~\ref{tab:ambrosia_unambig} shows, methods which disambiguate
first, show better results by a wide margin (compared to end-to-end
systems), with our proposed method achieving the best score
of~77.9\%. Explicit disambiguation serves as an intermediate
representation of the question, thereby clarifying its meaning.
Our results further demonstrate that generating a single
interpretation is less challenging than handling multiple valid
interpretations simultaneously (compare Table~\ref{tab:main_results}).

\begin{table}[t]
  \centering
  \begin{minipage}{0.47\linewidth}
  \raggedright
    \begin{tabular}{@{}lc@{}}
      \toprule
      \multicolumn{2}{@{}c@{}}{\textit{End-to-End}}\\  
      \midrule
      0-shot Prompt &  35.9 \\
      3-shot Prompt &  43.0 \\ 
      SFT  & 44.5 \\
      \bottomrule
    \end{tabular}
  \end{minipage}%
  \begin{minipage}{0.5\linewidth}
    \raggedleft
    \begin{tabular}{@{}lc@{}}
      \toprule
      \multicolumn{2}{@{}c@{}}{\textit{Disambiguate-and-Parse}}\\  
      \midrule
      Interp. Prompt & 75.1 \\
      Gold Interp. SFT &  56.9 \\ 
      Ours  &  \textbf{77.9} \\
      \bottomrule
    \end{tabular}
  \end{minipage}
  \caption{Coverage (\%) on \textbf{unambiguous} subset of Ambrosia test set
    (single coverage is the same as full coverage since each question
    has only one interpretation).}
  \label{tab:ambrosia_unambig}
\end{table}

\paragraph{The infilling module is robust to different interpretation types.}
We now evaluate our approach when the infilling model is trained on
Ambrosia. Specifically, we re-split Ambrosia to use
80\% for training (1K ambiguous and 2K unambiguous examples), 10\% for
validation, and 10\% for testing (131 ambiguous examples). All
databases in the test set are unseen during training, similar to the
original split.  Note that in this setting the infilling model is
trained on human-written interpretations which Ambrosia provides. As a
result, Ambrosia becomes the in-domain test set, whereas AmbiQT
represents out-of-domain evaluation.

Table~\ref{tab:ambrosia_train} shows results for an end-to-end
fine-tuned text-to-SQL model, and three variants of the disambiguate
first parse later framework: a model fine-tuned on gold (Ambrosia)
interpretations, a prompt-based model that generates default
interpretations without infilling, and the full model with
infilling. The latter achieves the best single and full
interpretation coverage on the out-of-domain AmbiQT test set and the
highest single interpretation coverage on Ambrosia. However, the full
interpretation coverage on Ambrosia is much lower than the end-to-end
fine-tuned baseline. Manual examination of the  predicted
interpretations revealed that some correct interpretations are parsed
incorrectly during the text-to-SQL stage, which relies on zero-shot
prompting and may not capture dataset-specific conventions. This
finding is supported by the results of the model fine-tuned on
interpretations, which also performs significantly worse than
end-to-end fine-tuning.  Finally, we found that training on both
datasets leads to results similar to the in-domain setting, see
Appendix~\ref{appx:additional_results} for details.

\begin{table}[t]
  \centering
\begin{tabular}{@{}l@{\;\;}c@{\;\;\;}cc@{\;\;\;}c@{}}
\toprule
& \multicolumn{2}{c}{AmbiQT} & \multicolumn{2}{c}{Ambrosia*} \\
\cmidrule(lr){2-3} \cmidrule(l){4-5}
Method  & Single & Full & Single & Full \\
\midrule
Text-to-SQL FT &  56.1 & 10.2 & 77.1 & \textbf{66.4}  \\
Interp. Prompt & 81.8 & 26.0 & 81.9 & 16.9 \\
Gold Interp. SFT & 81.9  & 4.1  & 71.0 & 38.9 \\ 
Ours  &  \textbf{88.0} &  \textbf{30.0}  & \textbf{87.8} & 30.5 \\
\bottomrule
\end{tabular}
\caption{\textbf{Single} and \textbf{Full} Interpretation Coverage (\%) on AmbiQT and
  Ambrosia* test sets. Symbol~* denotes our split of the Ambrosia test
  set into training, development, and testing. Models are fine-tuned on
  Ambrosia* train.}
\label{tab:ambrosia_train}
\end{table}

\section{Related Work}
 \paragraph{Ambiguity Resolution in NLP} Numerous studies have focused on ambiguity in natural language tasks using strategies like generating multiple answers \citep{min-etal-2020-ambigqa}, asking clarification questions \citep{lee-etal-2023-asking, Zhang2024ModelingFC}, and estimating uncertainty \citep{cole-etal-2023-selectively}. Similar to our work, \citet{sun-etal-2023-answering} use iterative prompting to refine and generate alternative interpretations in the context of question answering, while \citet{kim-etal-2024-aligning} first detect ambiguous questions and then resolve them through clarification requests. 

 \paragraph{Ambiguity in Semantic Parsing} Ambiguity has been studied across semantic parsing tasks, from code generation \citep{li-etal-2023-python, mu2023clarifygpt} to $\lambda$-calculus translation \citep{rasmussen-schuler-2020-corpus},  and logical form prediction \citep{eskin-2024-zero}. In the domain of text-to-SQL parsing, recent work has emphasized the fact that benchmarks often overlook ambiguity by providing single interpretations \citep{Floratou2024, pourreza-rafiei-2023-evaluating}. Existing approaches focus on detecting column ambiguity through counterfactual examples \cite{wang-etal-2023-know}, special-purpose decoding \citep{bhaskar-etal-2023-benchmarking}, and resolving ambiguity through clarification questions \citep{practiq-dataset}. Our work uses explicit disambiguation before parsing and thus extends to different types of ambiguities, question styles, and database formats.

\paragraph{Intermediate Representations in Text-to-SQL} Intermediate representations are commonly used to bridge the gap between natural language and database queries. Several approaches decompose complex questions into a sequence of simpler operations expressed in natural language   \cite{wolfson-etal-2022-weakly,saparina-osokin-2021-sparqling}, modular execution plans \cite{eyal-etal-2023-semantic}, or simplify the parsing task by augmenting questions with SQL keywords that mention necessary computation steps \cite{liu2024keyinstkeywordinstructionimproving, caferoğlu2024esqldirectschemalinking}. Building on this work, we also use intermediate representations to make implicit information explicit but focus on resolving ambiguity.

\paragraph{Learning to Correct LLM Outputs} More recently, various approaches have been proposed to correct systematic biases in LLM outputs. For example, \citet{ji2024alignerefficientalignmentlearning} propose a model-agnostic module that learns correctional residuals between preferred and dispreferred outputs. \citet{welleck2023generating} use a corrector model to iteratively review imperfect generations from a base model. 
Similarly,  critique generators can be developed using reinforcement learning  \cite{wadhwa-etal-2024-learning-refine}  or through fine-grained feedback \cite{wadhwa-etal-2024-learning}. Such correction approaches are most effective when guided by external tools \cite{kamoi-etal-2024-llms}. We follow this paradigm using a specialized infilling model to correct systematic  LLM biases towards certain interpretations and validate our output through SQL execution.

\section{Conclusion}
In this work, we present a novel approach for handling ambiguity in text-to-SQL semantic parsing. We first disambiguate questions by explicitly verbalising their interpretations and then leverage LLM capabilities to predict all valid SQL queries. A generator (LLM)  provides an initial set of default interpretations, which are then augmented with a specialized infilling model. We propose a method for training this model based on automatic annotations obtained by comparing SQL query execution results rather than natural language interpretations.

Our experimental results on AmbiQT and Ambrosia demonstrate the effectiveness of our approach. Our method achieves the highest single interpretation coverage on both datasets and maintains consistent full interpretation coverage in both in-domain and out-of-domain evaluation. However, generating all valid interpretations remains challenging. Future work could explore ways to improve ambiguity handling through better processing of database structure or using query execution as a signal for both training and test-time search.

\section{Limitations}
Our approach relies on reference interpretations and SQL queries for training the infilling model. While we show how to generate synthetic interpretations, scaling to new domains might be challenging. The sequential pipeline can propagate errors, with mistakes in disambiguation affecting text-to-SQL parsing and parser errors leading to incorrect SQL predictions. Duplicates in generated interpretations can help capture valid meanings through different wording but at the cost of longer output. 
The generator can produce incorrect interpretations that are currently not filtered. The infilling model can also miss valid interpretations that are underrepresented in the training data. 

Additionally, generating and parsing multiple interpretations requires more computation than single-stage approaches.
Although our approach handles both ambiguous and unambiguous questions, it occasionally provides multiple interpretations for unambiguous requests. Future work should focus on improving precision through better filtering and interpretation validation.

\section{Acknowledgements}
We thank the meta-reviewer and anonymous reviewers for their constructive feedback. We gratefully acknowledge the
support of the UK Engineering and Physical Sciences Research Council (grant EP/W002876/1).

\bibliography{anthology, custom}

\newpage
\appendix

\section{AmbiQT and Ambrosia Examples}\label{appx:dataset_examples}
We provide samples from the AmbiQT \cite{bhaskar-etal-2023-benchmarking} 
and Ambrosia \cite{Ambrosia} 
evaluation sets.

\begin{table}[h!]
\centering
\footnotesize
\settowidth\tymin{\textbf{Database Schema  Schema}}
\begin{tabulary}{\linewidth}{@{}L@{\;}L@{}}
\toprule
\multicolumn{2}{c}{\textbf{\normalsize{AmbiQT Examples}}}\\\midrule
\multicolumn{2}{c}{\textbf{Column Ambiguity}}\\\midrule%
\textbf{Database} & 
 \texttt{singer: singer\_id, \textbf{artist\_name}, \textbf{performer\_name}, song\_name, age, country  $\dots$} \\\addlinespace[3pt]
\textbf{Question} & Show \textbf{name}, country, age for all singers ordered by age from the oldest to the youngest. \\\addlinespace[3pt]
\textbf{SQL Query 1 } & \texttt{SELECT \textbf{artist\_name}, country, age FROM singer ORDER BY age DESC}\\\addlinespace[3pt]
\textbf{SQL Query 2 } & \texttt{SELECT \textbf{performer\_name}, country, age FROM singer ORDER BY age DESC}\\\midrule
\multicolumn{2}{c}{\textbf{Table Ambiguity}}\\\midrule%
\textbf{Database} & 
 \texttt{\textbf{Canine\_Breeds}: breed\_name, breed\_code}\\
 & \texttt{\textbf{Dog\_Types}:  breed\_name, breed\_code} \\
 & \texttt{dogs :  owner\_id, breed\_code $\dots$} \\\addlinespace[3pt]
\textbf{Question} & What is the name of the \textbf{breed} with the most dogs? \\\addlinespace[3pt]
\textbf{SQL Query 1 } & \texttt{SELECT T1.breed\_name \newline
 FROM \textbf{Canine\_Breeds} T1  JOIN dogs T2 $\dots$  ORDER BY COUNT(*) DESC LIMIT 1}
\\\addlinespace[3pt]
\textbf{SQL Query 2 } &  \texttt{SELECT T1.breed\_name \newline FROM \textbf{Dog\_Types} T1  JOIN dogs T2 $\dots$ \newline ORDER BY COUNT(*) DESC LIMIT 1}
\\\midrule
\multicolumn{2}{c}{\textbf{Join Ambiguity}}\\\midrule%
\textbf{Database} & 
 \texttt{\textbf{country}: surfacearea, indepyear, name, population, code $\dots$}\\
 & \texttt{\textbf{country\_surfacearea}:   surfacearea , code $\dots$} \\\addlinespace[3pt]
\textbf{Question} & What are the name, independence year, and \textbf{surface area of the country} with the smallest population? \\\addlinespace[3pt]
\textbf{SQL Query 1 } & \texttt{SELECT T1.name, T2.surfacearea, T1.indepyear \textbf{FROM country T1 }\newline  \textbf{JOIN country\_surfacearea T2} \newline\textbf{ ON T1.code = T2.code} \newline ORDER BY population LIMIT 1}
\\\addlinespace[3pt]
\textbf{SQL Query 2 } &  \texttt{SELECT name, surfacearea, indepyear FROM \textbf{country} \newline ORDER BY population LIMIT 1}
\\\midrule
\multicolumn{2}{c}{\textbf{Precomputed Aggregates}}\\\midrule%
\textbf{Database} & 
 \texttt{\textbf{show}: result, \textbf{attendance}, show\_id $\dots$}\\
 & \texttt{\textbf{show\_attendance}:   \textbf{avg\_attendance}, sum\_attendance $\dots$} \\\addlinespace[3pt]
\textbf{Question} & What is the \textbf{average attendance} of shows?\\\addlinespace[3pt]
\textbf{SQL Query 1 } & \texttt{SELECT \textbf{AVG(}attendance\textbf{)} \newline  FROM show}
\\\addlinespace[3pt]
\textbf{SQL Query 2 } &  \texttt{SELECT \textbf{avg\_attendance} \newline FROM show\_attendance}
\\\bottomrule
\end{tabulary}
\end{table}
\newpage
\begin{table}[t!]
\centering
\footnotesize
\settowidth\tymin{\textbf{Database Schema  Schema}}
\begin{tabulary}{\linewidth}{@{}L@{\;}L@{}}
\toprule
\multicolumn{2}{c}{\textbf{\normalsize{Ambrosia Examples}}}\\\midrule
\multicolumn{2}{c}{\textbf{Scope Ambiguity}}\\\midrule
\textbf{Database} & 
 \texttt{\textbf{Spa}: spa\_id, name, address \newline
  \textbf{Treatments}: treatment\_id, name  $\dots$ \newline
 Spa\_Treatments: spa\_treatment\_id, spa\_id, treatment\_id 
} \\\addlinespace[3pt]
\textbf{Question} & What treatment options do we have \textbf{for each spa?} \\\addlinespace[3pt]
\textbf{SQL Query 1 } & \texttt{SELECT T.name  FROM Spa\_Treatments ST \newline  JOIN Treatments T $\dots$ JOIN Spa S $\dots$   \newline 
\textbf{GROUP BY ST.treatment\_id HAVING   \newline COUNT(DISTINCT ST.spa\_id) = \newline 
(SELECT COUNT(*) FROM  Spa)}}
\\\addlinespace[3pt]
\textbf{SQL Query 2 } &  \texttt{SELECT \textbf{S.name, T.name}  \newline FROM Spa\_Treatments ST \newline  JOIN Treatments T $\dots$ JOIN Spa S $\dots$  }
\\\midrule
\multicolumn{2}{c}{\textbf{Attachment Ambiguity}}\\\midrule
\textbf{Database} & 
 \texttt{EventSpaces: id, Name, \textbf{Event\_Space}, Capacity, Address, ContactInfo $\dots$} \\\addlinespace[3pt]
\textbf{Question} & List all banquet halls and conference rooms \textbf{with a 200 person capacity}. \\\addlinespace[3pt]
\textbf{SQL Query 1 } & \texttt{SELECT ES.Name FROM EventSpaces ES} \newline  \texttt{WHERE } \newline \texttt{\textbf{(}ES.Event\_Space =
"Banquet Hall"} \newline \texttt{OR ES.Event\_Space = "Conference Room"\textbf{) }\newline\texttt{AND ES.Capacity = 200}
}
\\\addlinespace[3pt]
\textbf{SQL Query 2 } &  \texttt{SELECT ES.Name FROM EventSpaces ES} \newline  \texttt{WHERE} \newline \texttt{ES.Event\_Space =
"Banquet Hall"} \newline \texttt{\textbf{OR} ES.Event\_Space = "Conference Room" }\newline\texttt{\textbf{AND} ES.Capacity = 200}\\\midrule
\multicolumn{2}{c}{\textbf{Vagueness}}\\\midrule
\textbf{Database} & 
 \texttt{hospitals: id, name, \textbf{city}, \textbf{neighborhood}, phone\_number $\dots$} \\\addlinespace[3pt]
\textbf{Question} & \textbf{Where} are the clinics located? \\\addlinespace[3pt]
\textbf{SQL Query 1 } & \texttt{SELECT \textbf{neighborhood} \newline FROM hospitals \newline WHERE name LIKE '\%Clinic\%'}\\\addlinespace[3pt]
\textbf{SQL Query 2 } & \texttt{SELECT \textbf{city} \newline  FROM hospitals \newline  WHERE name LIKE '\%Clinic\%'} \\\addlinespace[3pt]
\textbf{SQL Query 3 } & \texttt{SELECT \textbf{neighborhood}, \textbf{city} \newline 
 FROM hospitals \newline  WHERE name LIKE '\%Clinic\%'}
\\\bottomrule
\end{tabulary}
\end{table}

AmbiQT contains column and table ambiguities, join ambiguity, and ambiguity due to precomputed aggregates. Ambrosia covers scope and attachment ambiguities, and vagueness.
Note that column and table ambiguities from AmbiQT correspond to vagueness in Ambrosia, although they differ in the number of gold queries provided (2 in AmbiQT versus 3 in Ambrosia).

AmbiQT is publicly available under the MIT license and Ambrosia is under the
CC BY 4.0 license.

\newpage

\section{Prompt for Annotating AmbiQT with Interpretations}\label{appx:AmbiQT_syn_prompt}
We use the following prompt with three in-context examples to generate natural language interpretations for ambiguous questions in AmbiQT:

\begin{framed}
\vspace{-3mm}
\begin{Verbatim}[breaklines=true, 
                 breakanywhere=true,
                 breaksymbolright={},
                 breaksymbolsepleft=0pt,
                 breaksymbolindent=0pt,
                 breaksymbolleft={},
                 fontsize=\small]
Your task is to rewrite the question using a given word or phrase.

Examples:
Question: Show titles of songs and names of singers.
Please rewrite using "stage name":
Give me titles of songs and stage names of singers.

Question: Show the name of the conductor that has conducted the most number of orchestras.
Please rewrite using "director":
List the name of the director who has conducted the most number of orchestras.

Question: Return the id of the document with the fewest paragraphs.
Please rewrite using "passages":
What is the id of the document with the fewest passages?

Please provide rewritten question for the following instance. Do not add any explanation or description, output only the rewritten question.

Question:  ...
Please rewrite using ...
\end{Verbatim}
\vspace{-3mm}
\end{framed}

\section{Prompt for Default Interpretation Generation}\label{appx:default_interp_prompt}
The following prompt is used to generate default interpretations:
\begin{framed}
\vspace{-3mm}
\begin{Verbatim}[breaklines=true, 
                 breakanywhere=true,
                 breaksymbolright={},
                 breaksymbolsepleft=0pt,
                 breaksymbolindent=0pt,
                 breaksymbolleft={},
                 fontsize=\small]
You are tasked with analyzing questions and providing their possible interpretations. The questions are related to database queries and may be ambiguous or unambiguous.

Your task:
- List every distinct way the question could be understood
- Be thorough and consider all possible meanings
- Explore different ways the question could be interpreted
- Don't limit yourself to obvious interpretations

Important:
- List each interpretation on a separate line
- Do not include explanations or reasoning
- Focus on semantically different interpretations
- Be specific and precise in wording

Given the following database context:
...

Provide interpretations for this question:
...
\end{Verbatim}
\vspace{-3mm}
\end{framed}

\section{Interpretation Infilling Details}\label{appx:infill_details}
We use the following instructions for the model:
\begin{framed}
\vspace{-3mm}
\begin{Verbatim}[breaklines=true, 
                 breakanywhere=true,
                 breaksymbolright={},
                 breaksymbolsepleft=0pt,
                 breaksymbolindent=0pt,
                 breaksymbolleft={},
                 fontsize=\small]
The task is to review the provided context, question, and existing interpretations, and determine if any additional interpretations are missing. If there are missing interpretations, list them on separate lines without explanations. If all interpretations have already been covered, simply state: "All possible interpretations are covered."

Given the following context: ...

Question: ...

Existing interpretations: ...

Provide any missing interpretations or confirm that all possible interpretations are covered.
\end{Verbatim}
\vspace{-3mm}
\end{framed}

We fine-tune the instruction-tuned Llama 3.1 8B model using LoRA \citep{lora-hu-etal-2022}  with rank 16 ($\alpha=16$) and NEFTune \citep{jain2024neftune} (noise $\alpha=5$). The model is trained for 15 epochs using a cosine learning rate schedule with an initial learning rate of 5e-5, weight decay of 0.01, and a warmup ratio of~0.01. Training is performed on a single NVIDIA A100 GPU with a batch size of 8 and gradient clipping at 0.3. The total time for training and evaluation of one run is under 10 hours. We sample 10\% of the training data as development set 
and select the best-performing model from a single run for final evaluation.

We apply the same fine-tuning procedure to all comparison methods, i.e., end-to-end text-to-SQL fine-tuning and fine-tuning to predict interpretations. We observe that these methods tend to overfit with more epochs and thus reduce the number of training epochs to~5.

\section{Text-to-SQL Parsing}\label{appx:text2sql_prompt}
To generate SQL queries for unambiguous questions (or interpretations), we use the following prompt across all text-to-SQL tasks, including AmbiQT annotation validation, evaluation of baseline models with interpretations, and our approach:

\begin{framed}
\vspace{-3mm}
\begin{Verbatim}[breaklines=true, 
                 breakanywhere=true,
                 breaksymbolright={},
                 breaksymbolsepleft=0pt,
                 breaksymbolindent=0pt,
                 breaksymbolleft={},
                 fontsize=\small]
The task is to write SQL queries based on the provided questions in English. Questions can take the form of an instruction or command. Do not include any explanations, and do not select extra columns beyond those requested in the question.

Given the following SQLite database schema: ...

Answer the following: ...
\end{Verbatim}
\vspace{-3mm}
\end{framed}

\section{Self-Correction Prompt}\label{appx:self_correction}
We compare our approach against a self-correction method where the LLM is prompted to review the generated interpretations (Appendix \ref{appx:default_interp_prompt}), choose the valid ones, and add any missing interpretations. The prompt is shown below:

\begin{framed}
\vspace{-3mm}
\begin{Verbatim}[breaklines=true, 
                 breakanywhere=true,
                 breaksymbolright={},
                 breaksymbolsepleft=0pt,
                 breaksymbolindent=0pt,
                 breaksymbolleft={},
                 fontsize=\small]
The task is to review the provided context, question, and candidate interpretations, and based on this information provide the interpretations that accurately reflect the meaning (or one of the possible meanings) of the question. If any of the candidate interpretations are correct, provide them as a list of interpretations. If there are missing interpretations, provide them as well. Avoid providing interpretations that are incorrect or duplicates. Do not provide any explanations.

Given the following context: ...

Question: ...

Candidate interpretations: ...

Provide the interpretations that accurately reflect the meaning (or one of the possible meanings) of the question.
\end{Verbatim}
\vspace{-3mm}
\end{framed}

\section{Additional Results}\label{appx:additional_results}
\looseness=-1
\paragraph{Comparison with a larger-sized model} 
Table~\ref{tab:larger_baseline} extends Table~\ref{tab:main_results} by including a larger baseline: the instruction-tuned Llama 3.1 70B model. While the 70B variant outperforms the 8B version overall, our system (based on the 8B model) still achieves higher coverage and recall. We expect that larger models could further improve results, however, this comes at a significantly higher computational cost. %

\looseness=-1

\paragraph{Training on Ambrosia*} 
Table~\ref{tab:ambrosia_train_extended} extends Table~\ref{tab:ambrosia_train} with recall and precision. Consistent with earlier findings, our approach achieves the highest coverage and recall on the AmbiQT test set, demonstrating strong out-of-domain generalization. 

\begin{table*}[t]
\centering
\begin{tabular}{lcccccccc}
\toprule
 & \multicolumn{4}{c}{AmbiQT} & \multicolumn{4}{c}{Ambrosia} \\
\cmidrule(lr){2-5} \cmidrule(l){6-9}
\raisebox{.8em}[0pt]{\textit{Llama 3.1 70B}} & Single & Full & Recall & Precision & Single & Full & Recall & Precision \\
\midrule
0-shot Prompt         & 76.6     & 26.5  & 51.5   & 65.8   & 32.3     &  0.0  & 14.1   & 32.0   \\
3-shot Prompt         & 75.7     & 28.0  & 51.9   & \textbf{69.2}   & 57.5     &  3.9  & 28.1   & \textbf{50.9} \\ \bottomrule
\multicolumn{9}{c}{~}\\  \toprule
  & \multicolumn{4}{c}{AmbiQT} & \multicolumn{4}{c}{Ambrosia} \\
\cmidrule(lr){2-5} \cmidrule(l){6-9}
\raisebox{.8em}[0pt]{\textit{Llama 3.1 8B}} & Single & Full & Recall & Precision & Single & Full & Recall & Precision \\\midrule
0-shot Prompt         & 62.3  & 12.3  & 37.3   & 58.1   & 29.4  &  0.9  & 15.0   & 21.9   \\
3-shot Prompt         & 44.3  & 10.9  & 27.6      & 33.0      & 35.7  &  1.3  & 17.5      & 21.3      \\ 
Ours                  & \textbf{92.3} & \textbf{53.2}  & \textbf{72.8}   & 38.2   & \textbf{84.4} & \textbf{18.8} & \textbf{51.9} & 24.2   \\
\bottomrule
\end{tabular}
\caption{Comparison of Llama 3.1 70B, Llama 3.1 8B and our approach  on AmbiQT and Ambrosia datasets. Our approach is based on Llama 3.1 8B and \textbf{fine-tuned on AmbiQT train}.
}
\label{tab:larger_baseline}
\end{table*}

\begin{table*}[h!]
\centering
\begin{tabular}{@{}lcccccccc@{}}
\toprule
& \multicolumn{4}{c}{AmbiQT} & \multicolumn{4}{c}{Ambrosia*} \\
\cmidrule(lr){2-5} \cmidrule(l){6-9}
Method & Single & Full & Recall & Precision & Single & Full & Recall & Precision \\
\midrule
Text-to-SQL FT     & 56.1 & 10.2 & 33.2 & 52.7 & 77.1 & \textbf{66.4} & \textbf{71.6} & \textbf{73.3} \\
Interp.\ Prompt    & 81.8 & 26.0 & 53.4 & 31.8 & 81.9 & 16.9 & 49.0 & 29.1 \\
Gold Interp.\ SFT  & 81.9 & 4.1  & 43.0 & \textbf{79.5} & 71.0 & 38.9 & 53.7 & 61.5 \\
Ours               & \textbf{88.0} & \textbf{30.0} & \textbf{58.5} & 35.3 & \textbf{87.8} & 30.5 & 59.7 & 26.3 \\
\bottomrule
\end{tabular}
\caption{
\textbf{Single} and \textbf{Full} Interpretation Coverage, \textbf{Recall} and \textbf{Precision} (\%) on AmbiQT and Ambrosia* test sets. Models are \textbf{fine-tuned on Ambrosia* train}. Symbol~* denotes our split of the Ambrosia test set into training, development, and testing.
}
\label{tab:ambrosia_train_extended}
\end{table*}

\begin{table*}[h!]
\centering
\begin{tabular}{@{}lcccccccc@{}}
\toprule
& \multicolumn{4}{c}{AmbiQT} & \multicolumn{4}{c}{Ambrosia*} \\
\cmidrule(lr){2-5} \cmidrule(l){6-9}
Method & Single & Full & Recall & Precision & Single & Full & Recall & Precision \\
\midrule
Text-to-SQL FT     & 81.0 & \textbf{63.1} & 72.1 & 73.3 & 83.2 & \textbf{69.5} & \textbf{76.2} & \textbf{78.1} \\
Gold Interp.\ SFT  & 86.2 & 60.4 & \textbf{73.3} & \textbf{79.8} & 75.6 & 37.4 & 55.0 & 65.1 \\
Ours               & \textbf{92.5} & 54.0 & 73.2 & 38.8 & \textbf{84.7} & 29.8 & 57.1 & 28.8 \\
\bottomrule
\end{tabular}
\caption{
\textbf{Single} and \textbf{Full} Interpretation Coverage, \textbf{Recall} and \textbf{Precision} (\%) on AmbiQT and Ambrosia* test sets. All models are \textbf{fine-tuned on both AmbiQT and Ambrosia* train}. Symbol * denotes our split of the Ambrosia test set into training, development, and test.
}
\label{tab:full_train}
\end{table*}

\paragraph{Training on AmbiQT and Ambrosia*} Table~\ref{tab:full_train} provides results when the models are trained on the maximum available data: the AmbiQT train set and re-split Ambrosia* training set. As both test sets are in-domain, we would not expect to see any advantages from our method in terms of full coverage. Nevertheless, it still provides the best single-interpretation coverage. Overall, the results in Table~\ref{tab:full_train} are very similar to those obtained in the in-domain setting.

\end{document}